\documentclass{article}
\usepackage{spconf,amsmath,graphicx}
\usepackage{multirow}
\usepackage[utf8]{inputenc}
\usepackage{authblk}
\usepackage{tikz}

\title{HYPER-SPECTRAL IMAGING FOR OVERLAPPING PLASTIC FLAKES SEGMENTATION}
\makeatletter
\def\@name{\textit{Guillem Martinez$^{1}$, Maya Aghaei$^{1}$, Martin Dijkstra$^{1}$, Bhalaji Nagarajan$^{2}$, Femke Jaarsma$^{1}$,\\ Jaap van de Loosdrecht$^{1}$, Petia Radeva$^{2,3}$, Klaas Dijkstra$^{1}$}}
\makeatother

\address{(1) NHL Stenden University of Applied Sciences, Leeuwarden, The Netherlands,\\ 
(2) Dept. de Matemàtiques i Informàtica, Universitat de Barcelona, Barcelona, Spain, \\
(3) Computer Vision Center,  Cerdanyola del Vallés (Barcelona), Spain}
\begin{document}

\maketitle

%

%
\begin{abstract}
Given the hyper-spectral imaging unique potentials in grasping the polymer characteristics of different materials, it is commonly used in sorting procedures. In a practical plastic sorting scenario, multiple plastic flakes may overlap which depending on their characteristics, the overlap can be reflected in their spectral signature. In this work, we use hyper-spectral imaging for the segmentation of three types of plastic flakes and their possible overlapping combinations. We propose an intuitive and simple multi-label encoding approach, bitfield encoding, to account for the overlapping regions. With our experiments, we show that the bitfield encoding improves over the baseline single-label approach and we further demonstrate its potential in predicting multiple labels for overlapping classes even when the model is only trained with non-overlapping classes.
\end{abstract}
\begin{keywords}
Hyper-spectral imaging, plastic sorting, multi-label segmentation, bitfield encoding
\end{keywords}
\section{Introduction}
\label{sec:intro}
Plastics are part of our daily routine, which despite their advantages have tremendous environmental disadvantages including the required resources for production and toxicity \cite{toniolo2013comparative}. Plastic recycling hence is crucial to mitigate this negative environmental footprint. Within the recycling procedure, plastic objects are collected, shredded into flakes, and washed as a preparation for analysis of their polymer type \cite{voet2021plastics}. In this work, we focus on Short-Wave Infrared (SWIR) Hyper-Spectral Imaging (HSI) for polymer types analysis of the previously shredded plastic flakes. Due to the unique absorption of the electromagnetic spectrum by different polymers (i.e., the polymer spectral signature), differentiation between them using HSI becomes possible \cite{serranti2019plastic}. This method does not require sample preparation and can be done in near real-time, which is an advantage over typical analytical methods \cite{eldin2011near}. 

\begin{figure}
\centering
	\includegraphics[width=0.5\textwidth]{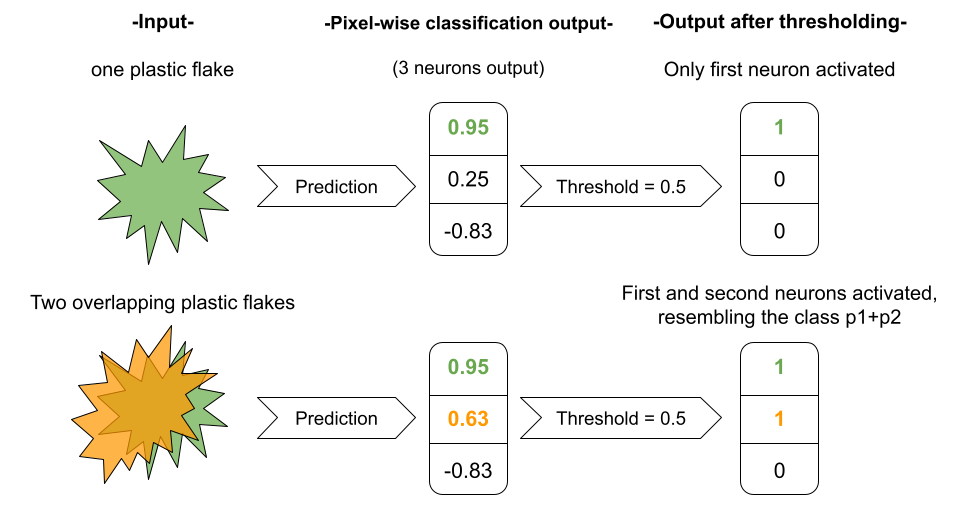}
	\caption{Bitfield-encoding depiction for pixel-wise classification of plastic flakes, given in two scenarios: a single flake and two overlapping flakes.}
	\label{fig:bitfield}
\end{figure}

In a practical recycling scenario, using HSI, mixed plastic flakes directly from the shredder are randomly placed on the conveyor belt to pass under a Hyper-Spectral (HS) camera. In this process, two or more flakes may physically overlap each other which leads to a disturbed spectral signature perception. We recognized this issue and propose a simple yet effective method to recognize the overlapped classes of plastics, given only the primary (non-overlapped) classes during training. Single-label to multi-label problem is considered an important problem in computer vision and image processing literature \cite{wang2016cnn}. In this work, we propose to encode the per-pixel label as a vector of single binary bits instead of the commonly used one-hot encoding and refer to this as bitfield encoding. To put this into perspective, assuming that each single bit in the bitfield encoding data-structure corresponds to a primary plastic class in our dataset, the overlap of two or more classes of plastics can be identified as activating the corresponding bits to each of the individual overlapping classes. In this way, we can represent the overlap of classes with the same number of bits, but activating more than one bit at the same time (see Fig. \ref{fig:bitfield}). We further demonstrate that this representation not only allows a more intuitive interpretation of multi-label classification problem but also enables multi-label classification providing only single-label data in training. Bitfield encoding can be applied with any classification technique,  however, in this work, we mainly focus on its application with pixel-wise classification using the U-net segmentation framework \cite{ronneberger2015u}.


\section{State of the art}
\label{sec:sota}

Multi-fold developments of deep learning enabled vast improvements in HSI \cite{ozdemir2020deep}, leading to the emergence of spectral or spectral-spatial architectures \cite{sun2021supervised}. HybridSN \cite{roy2019hybridsn} used a combination of spectral-spatial 3D CNN and spatial 2D CNN. Squeeze-and-Excitation (SE) residual bag-of-feature learning framework used SE blocks and batch normalization to suppress feature maps that are irrelevant to the learning process. SpectralNET \cite{chakraborty2021spectralnet} used wavelet CNNs instead of a 3D CNN to learn the spectral features as they are computationally lighter compared to a 3D CNN computation. The recent advances in deep learning - attention mechanisms and transformers are also documented in HSI. HSI-BERT \cite{he2019hsi} used a multi-head self-attention mechanism to encode the semantic context-aware representations in order to obtain discriminatory features and A2S2K-ResNet \cite{roy2020attention} used 3D ResBlocks and Explainability Failure Ratio (EFR) mechanisms to learn the spectral-spatial features for HSI classification.

Among the vast number of applications, plastic sorting is an active application of HSI \cite{araujo2021review}. In any sorting scenario, it is not always possible to present distinct samples to the imaging system. Real-world samples are often multi-labelled in nature, where the samples are overlapping each other. Multi-label classification methods hence are proposed to solve such problems with either problem transformation or algorithm adaptation techniques \cite{tsoumakas2009correlation}. Problem transformation methods convert the problem into other learning scenarios such as Binary Relevance or Label Powerset. Algorithm adaptation methods use the multi-label data directly. The algorithm adaptation methods suffer from label concurrence and often encounter imbalanced label sets \cite{tarekegn2021review}. Ensemble methods \cite{wang2021active} are effective as a problem transformation method, where each network is trained  with a random k-label set treating each as a single label task. However, they are computationally complex and model selection criterion is not well defined. Label-specific features \cite{fang2021multi} have been studied to learn discriminative features for each label. 
One of the systematic errors in large deep learning datasets is where the samples are labeled as single-labeled but are multi-labeled in nature \cite{yun2021re}. Iterated learning frameworks \cite{rajeswar2021multi} have been widely studied to tackle this labeling bias, however, it suffers from higher training periods. Re-labeling strategies \cite{yun2021re} offer a more cleaner solution. The main consideration in these algorithms is the additional cost and the training duration.

HSI datasets are usually small in size, sometimes containing few to even only one HS image. Thus, deep learning algorithms satisfy their need for large datasets using data augmentation strategies, such as those used in Pixel-block pair \cite{li2018data} or using GANs for HSI \cite{roy2021generative}. With the above discussed algorithmic limitations and considerations in existing literature, we propose a new bitfield encoding, through defining single and overlapping classes directly using a vector of single bits, thereby reducing the complexity of the learning algorithm. In addition, we introduce a new dataset containing overlapping as well as non-overlapping plastics, composed of HS images of three primary plastics and their possible overlapping combinations. This dataset is acquired by capturing the HS signature of plastic flakes randomly scattered over a conveyor belt passing under the Specim FX17 camera\footnote{https://www.specim.fi/products/specim-fx17/}. 


\section{Bitfield encoding for overlapped plastic segmentation}
\label{sec:method}

Commonly, image segmentation requires a label for each pixel of the image.  To use this approach for plastic HSI, the label must include classes for overlaps of plastics for further recognition of them. A limitation with this approach is that the number of classes for the combination of polymers grows exponentially as the number of primary polymers increases. In the same way, obtaining enough data for balanced analysis of overlapped polymers with regards to the primary polymers is also non viable. Given the HS signature of each plastic individually, we aim to answer if it is possible to recognize when two or more classes (polymers) are overlapping, given only the non-overlapping classes (polymers) when training the models.

We propose an intuitive approach to this problem, namely bitfield encoding: instead of labeling each pixel with a single class, we propose to label each pixel with a bitfield vector, where every bit corresponds to one primary polymer. Note that a vector of three bits provides eight different combinations. With this encoding, pixels corresponding to the overlap of two or more polymers can be annotated with activating in the vector, all the bits corresponding to each of the overlapping polymers. This approach enables to implicitly pass the knowledge that some classes (the overlapped pixels) are a combination of primary ones. 


In this work, we propose to use bitfield encoding with U-net \cite{ronneberger2015u} segmentation framework. For HSI analysis in this study, we adapted the original U-net structure to accept input size H $\times$ W $\times$ 224 HS data. At the output level, Mean Squared Error (MSE) is used as loss function together with Hyperbolic Tangent (TanH) as activation function. TanH being 0-centered, leads to the faint classes being more likely to appear at the output after thresholding. The shape at the output is $H$ $\times W \times K$, where $K$ (in our case, equal to 3) is the number of primary classes, also representing the size of the bitfield vector. A threshold parameter is set to select the bits with the highest probability. An example of a prediction using bitfield encoding is given in Fig.~\ref{fig:bitfield}. In the figure, note that the prediction can have values in between $\left[ -1,1 \right]$, given the TanH activation function. 

The adapted U-net output for each primary class is given for the scenarios when a single flake or overlapping of them is scanned. After thresholding at 0.5, in the single flake scenario, only one bit corresponding to the predicted plastic is activated. In the case of overlapping plastics, we achieve the bitfield encoding representing the overlap of plastics 1 and 2, as both these bits have scored over 0.5.



\begin{figure}
\centering
	\includegraphics[height=0.25\textwidth]{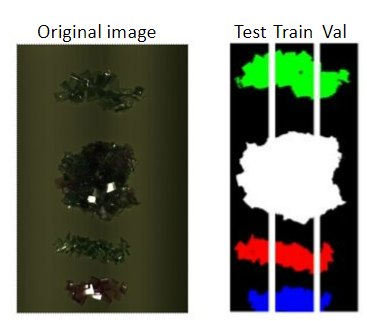}
	\caption{Left: RGB representation of the original HS image with the total number of four blobs. Each slice of an original image is divided into testing, training and validation subsets from left to right. Each color represents a different label for the blob, including a blob of overlapping plastics.}
	\label{f:vs-dataset}
\end{figure}

\section{Experiments and Results} \label{sec:experiments}

\subsection{HS Overlaps Dataset}
\label{sec:dataset}

The HS Overlaps Dataset (HSOD) contains 18 images, each containing blobs of plastics from three different types, with and without overlap. The original images have a resolution of 996 $\times$ 640 $\times$ 224, where the first two dimensions indicate the spatial resolution and  224 is the spectral resolution. The segmentation masks are automatically annotated using a combination of blob detection and morphological operations, introducing slight imprecision in the overlapping areas\footnote{The dataset and annotations will be publicly available with this paper.}. HSOD contains the following classes: background, three primary plastics (PP, PE, PET) and their four possible combinations (PP+PE, PP+PET, PE+PET, PP+PE+PET), resulting in eight possible classes.

Due to the sparsity of the available HS data in public datasets, (random) pixel selection from a single HS image to define training and evaluation subsets is a common practice \cite{lee2017going}. In the same line, to account for the class imbalance in HSOD (see the left-most column in Table \ref{table:results}), we opted for vertically dividing each original image by first removing margins from the background and then slicing the image into three equal parts of size 876 $\times$ 128 $\times$ 244. Each part is further used for either train, validation or test purposes (see Fig. \ref{f:vs-dataset}).  


\begin{table*}[]
\begin{tabular}{ccl|ccc|ccc|ccc}
\hline
\multicolumn{1}{l|}{\multirow{2}{*}{\# Blobs}} & \multicolumn{1}{l|}{\multirow{2}{*}{Encoding}} & \multicolumn{1}{c|}{\multirow{2}{*}{Category}} & \multicolumn{3}{c|}{Baseline}      & \multicolumn{3}{c|}{Baseline-Bitfield}   & \multicolumn{3}{c}{Bitfield} \\ \cline{4-12} 
\multicolumn{1}{l|}{}                           & \multicolumn{1}{l|}{}                          & \multicolumn{1}{c|}{}                          & F1    & Precision & Recall         & F1             & Precision      & Recall & F1     & Precision  & Recall \\ \hline
\multicolumn{1}{c|}{-}                          & \multicolumn{1}{c|}{000}                       & Background                                     & 0.998 & 0.998     & 0.998          & 0.998          & 0.999          & 0.996  & 0.992  & 1          & 0.985  \\
\multicolumn{1}{c|}{8}                         & \multicolumn{1}{c|}{001}                       & PP                                             & 0.982 & 0.972     & 0.992          & 0.979          & 0.961          & 0.997  & 0.553  & 0.388      & 0.964  \\
\multicolumn{1}{c|}{8}                         & \multicolumn{1}{c|}{010}                       & PE                                             & 0.969 & 0.968     & 0.969          & 0.949          & 0.914          & 0.987  & 0.741  & 0.604      & 0.960  \\
\multicolumn{1}{c|}{9}                         & \multicolumn{1}{c|}{100}                       & PET                                            & 0.942 & 0.898     & 0.990          & 0.963          & 0.939          & 0.989  & 0.430  & 0.390      & 0.481  \\
\multicolumn{1}{c|}{2}                          & \multicolumn{1}{c|}{011}                       & PP+PE                                          & 0.961 & 0.940     & 0.984          & 0.976          & 0.967          & 0.985  & 0.088  & 0.686      & 0.047  \\
\multicolumn{1}{c|}{3}                          & \multicolumn{1}{c|}{101}                       & PP+PET                                         & 0.923 & 0.986     & 0.868          & 0.940          & 0.982          & 0.902  & 0.340  & 0.278      & 0.447  \\
\multicolumn{1}{c|}{3}                          & \multicolumn{1}{c|}{110}                       & PE+PET                                         & 0.825 & 0.745     & 0.924          & 0.817          & 0.703          & 0.977  & 0.421  & 0.294      & 0.741  \\
\multicolumn{1}{c|}{3}                          & \multicolumn{1}{c|}{111}                       & PP+PE+PET                                      & 0.903 & 0.981     & 0.837          & 0.903          & 0.981          & 0.837  & 0.110  & 0.447      & 0.062  \\ \hline
\multicolumn{1}{l}{}                            & \multicolumn{1}{l}{}                           & Average                                        & 0.938 & 0.936     & \textbf{0.945} & \textbf{0.941} & \textbf{0.958} & 0.930  & 0.425  & 0.481      & 0.549  \\ \hline
\end{tabular}

\caption{HSOD statistics as well as the results of different experiments with it.}
\label{table:results}
\end{table*}

\subsection{Experiments}
We realized three experiments to incrementally demonstrate the potentials of bitfield-encoding. For all the experiments the testing and validation sets are intact, while the training set changes according to the experiment.

\begin{figure}[h!]
\centering
	\includegraphics[height=0.4\textwidth]{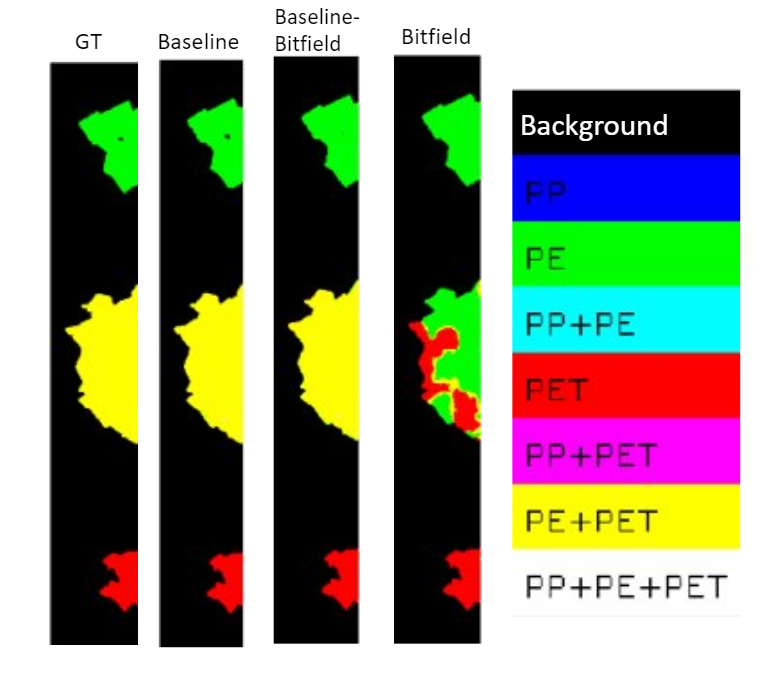}
	\caption{From left to right: Ground Truth, Baseline, Baseline-Bitfield, Bitfield.}
	\label{fig:results}
\end{figure}

\textbf{Baseline}: This experiment is designed to evaluate the performance of the baseline U-net model without any modification to the architecture: Cross-Entropy is used as the loss function and Softmax as the final activation function. The last layer contains eight neurons, one corresponding to each class in the dataset. Quantitative results of this experiment are given in Table \ref{table:results}-Baseline. In this experiment, we use the  primary plastics as well as overlapped plastics data for training the model. 


\textbf{Baseline-Bitfield}:
\textit{Does bitfield encoding enhance multi-class segmentation problems?} In this experiment, we still use the  primary and overlapped plastics data for training the model. However, to observe the difference in performance with regards to the Baseline, we modify the output layer to use MSE-Loss, Hyperbolic tangent, and at the output with three neurons. Quantitative results of this experiment are given in Table \ref{table:results}-Baseline-Bitfield.


\textbf{Bitfield}: In this experiment, the objective is to define the possibility of using bitfield encoding in classifying overlapped pixels when the model is only trained with primary plastics. We use additional HS blobs of primary plastics for training purposes (8 PP, 8 PE, 7 PET). The same output architecture as Baseline-Bitfield experiment is used for this experiment. Quantitative results of this experiment are given in Table \ref{table:results}-Bitfield.


As can be seen in Table \ref{table:results}, the Bitfield experiment shows poor performance compared to the Baseline and Baseline-Bitfield. However, the recall from the Bitfield experiment shows that the model is properly capable of detecting primary classes. The rather poor performance of the Bitfield experiment with regards to the Baseline experiments can also be seen in Fig. \ref{fig:results}. We can observe that the Baseline and the Baseline-Bitfield are almost identical to the ground truth. We can also verify that the Bitfield experiment trained with only primary classes predicts the primary polymer samples ideally, but is not powerful in predicting combined polymer samples. An interesting observation however is that the Bitfield model although not capable of detecting the overlap of pixels, is able to detect the correct plastics composing the overlap area (PE and PET instead of PE+PET).

\begin{figure}[h!]
\centering
	\includegraphics[height=0.35\textwidth]{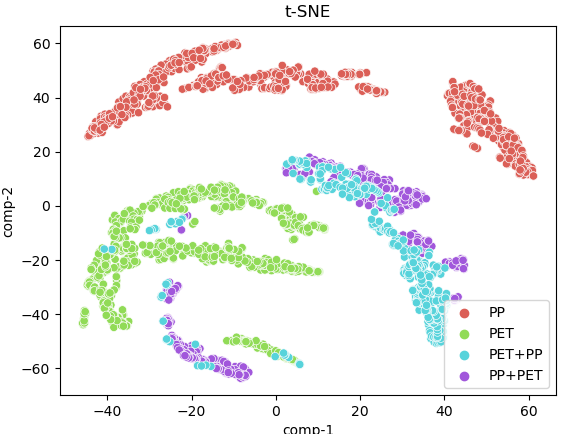}
	\caption{T-SNE representations of two primary plastics and their two-way overlapping order. The overlapping tends to locate in between the primary classes. The two overlapping representations are similar to each other, while separated from the representation of their primary classes.}
	\label{fig:tsne}
\end{figure}

In Fig. \ref{fig:tsne}, we show the T-SNE visualization of HS pixel values of four different blobs of samples: PP, PET, and their overlap in two ways, PP on top and PET on top. An interesting observation is that in this visualization, the overlapped classes appear in between the primary polymers and are fairly separated from the primary classes. However, according to this visualization, the order of the overlaps does not highly matter for their recognition. It is still worth mentioning that in this study we did not use any pre-processing method such as spectral normalization. Such pre-processing steps can be used in the future for smoothing possible over(under)-exposed pixel values, to improve the classification performance.

\section{Conclusions}

In this paper, we introduced the bitfield encoding as an approach for multi-label classification problem where some of the classes are overlapping of primary, single-label classes. To demonstrate the effectiveness of this approach, we compared its performance with a baseline, where the usual single-label is used to train and evaluate the models and showed that the bitfield encoding outperforms the traditional approach. Additionally, we showed the functionality of the bitfield encoding for prediction of class combinations (overlapped plastics) when the model is only trained with single classes (primary plastics). The latter is important as wider range of applications such as multiple label generation for datasets \cite{roig2020unsupervised} can benefit from it. 




\bibliographystyle{IEEEbib}
\bibliography{biblio}

\end{document}